\documentclass[]{spie}  

\usepackage{amsmath,amsfonts,amssymb}
\usepackage{graphicx}
\usepackage[colorlinks=true, allcolors=blue]{hyperref}
\usepackage{amsmath}
\usepackage{gensymb}
\usepackage{subfigure}

\newcommand{\be}{\mathbf{e}}
\newcommand{\bx}{\mathbf{x}}
\newcommand{\mbb}{\mathbb}
\newcommand{\mcl}{\mathcal}

\newcommand{\bv}{\mathbf{v}}

\newcommand{\lp}{\left(}
\newcommand{\rp}{\right)}

\DeclareMathOperator*{\argmin}{arg\,min} 
\DeclareMathOperator*{\argmax}{arg\,max}

\title{Graph-based Active Learning for Semi-supervised Classification of SAR Data\thanks{\hspace{1mm} {\bf Source code:} \url{https://github.com/jwcalder/MSTAR-Active-Learning}}}

\author[a]{Kevin Miller}
\affil[a]{University of California, Los Angeles, Department of Mathematics, 520 Portola Plaza, Los Angeles, CA 90095, USA}

\author[b]{John Mauro}
\affil[b]{Loyola Marymount University, Department of Mathematics, 1 LMU Drive, Los Angeles, CA 90045, USA}

\author[c]{Jason Setiadi}
\affil[c]{University of Minnesota, School of Statistics, 313 Ford Hall, 224 Church Street SE, Minneapolis, MN 55455, USA}

\author[d]{Xoaquin Baca}
\affil[d]{Harvey Mudd College, Department of Computer Science, 201 Platt Blvd, Claremont, CA, 91711, USA}

\author[a]{Zhan Shi}

\author[e]{Jeff Calder}
\affil[e]{University of Minnesota, School of Mathematics, 538 Vincent Hall, 206 Church Street SE, Minneapolis, MN 55455, USA}

\author[a]{Andrea L. Bertozzi}

\authorinfo{Further author information: (Send correspondence to K.M.)\\K.M.: E-mail: millerk22@g.ucla.edu}

\begin{document} 
\maketitle

\begin{abstract}
We present a novel method for classification of Synthetic Aperture Radar (SAR) data by combining ideas from graph-based learning and neural network methods within an active learning framework.  Graph-based methods in machine learning are based on a similarity graph constructed from the data. When the data consists of raw images composed of scenes, extraneous information can make the classification task more difficult. In recent years, neural network methods have been shown to provide a promising framework for extracting patterns from SAR images. These methods, however, require ample training data to avoid overfitting. At the same time, such training data are often unavailable for applications of interest, such as automatic target recognition (ATR) and SAR data. We use a Convolutional Neural Network Variational Autoencoder (CNNVAE) to embed SAR data into a feature space, and then construct a similarity graph from the embedded data and apply graph-based semi-supervised learning techniques. The CNNVAE feature embedding and graph construction requires no labeled data, which reduces overfitting and improves the generalization performance of graph learning at low label rates. Furthermore, the method easily incorporates a human-in-the-loop for active learning in the data-labeling process. We present promising results and compare them to other standard machine learning methods on the Moving and Stationary Target Acquisition and Recognition (MSTAR) dataset for ATR with small amounts of labeled data.  
\end{abstract}


\section{INTRODUCTION}
\label{sec:intro}  

Synthetic Aperture Radar (SAR) utilizes the movement of an antenna over a distance from the target to capture finer resolution images than standard radar. There are both phase and amplitude components of the signal that in combination can provide more detailed information about the objects in the scene. Automatic target recognition (ATR) of SAR data seeks to classify the objects of interest in such SAR images. Hand-labeling images by human eye is an impractical and time-consuming task for large datasets.  This makes SAR very amenable to automated machine learning methods. 

Supervised machine learning algorithms, such as deep learning, rely on an abundance of labeled data to learn from. In many applications, labeling data can be quite costly, while unlabeled data is ubiquitous and easy to obtain. Semi-supervised learning (SSL) methods use both labeled and unlabeled data in the learning task and aim to achieve good quality results with far less labeled data than fully supervised methods. A common way to use the unlabeled data is through the construction of a similarity graph, which effectively leverages relations between unlabeled datapoints for dimension reduction and classification tasks. The similarity graph structure can be exploited with standard graph-based SSL techniques, such as label propagation \cite{zhu_semi-supervised_2003}, also called Laplace learning, which propagates labels \emph{smoothly} over the graph via a diffusion process involving the graph Laplacian. 

A successful application of graph-based learning to image classification hinges on constructing a high-quality graph that accurately encodes the similarities between datapoints, in this case SAR images, that are important for the classification task at hand, while ignoring or suppressing differences between images that are due to spurious noise or image acquisition artifacts. Since the raw pixel values are sensitive to noise, contrast, lighting, or small shifts or rotations that commonly corrupt image data, it is important to apply a feature transformation to the images before constructing a graph. Standard feature transformations include the Scale Invariant Feature Transformation (SIFT)\cite{lowe1999object}, the scattering transform \cite{bruna2013invariant}, or a pre-trained neural network \cite{simonyan2015very}. Several recent papers have successfully used variational autoencoders (VAEs) for unsupervised feature extraction in hyperspectral imagery \cite{mei2019unsupervised}, SAR imagery \cite{deng2017sar,chen2014sar}, and for constructing similarity graphs in graph-based learning \cite{calder_poisson_2020,calder2022hamilton}.  VAE feature learning is an unsupervised method that retains the power and flexibility of deep supervised learning, making it ideal for problems with limited amounts of data.

In addition to constructing a high quality graph and considering the amount of labeled data available to a classifier, the {\it choice} of training (labeled) points can significantly affect classifier performance.   Active learning \cite{settles_active_2012} is a branch of machine learning that judiciously selects a limited number of unlabeled datapoints to query for labels, with the aim of maximally improving the underlying SSL classifier's performance. In applications like ATR in SAR imagery, the chosen active learning query points are labeled by an oracle, or human in the loop, such as a domain expert. These query points are selected by optimizing an {\it acquisition function} over the discrete set of datapoints available in the unlabeled pool of data. Active learning can greatly increase the performance of classifiers at very low label rates, and minimize the cost of labeling data with a human-in-the-loop.

In this work we present a novel pipeline for combining graph-based semi-supervised learning and VAE-based feature extraction methods within an active learning framework to improve ATR in SAR imagery data.  In particular, we use a convolutional variational autoencoder (CNNVAE) to learn feature representations of SAR images. The feature representations are used to embed the SAR images into a feature space where Euclidean similarities are more meaningful for constructing a similarity graph. The CNNVAE feature embedding is completely unsupervised (i.e., requires no labeled data), so the method is compatible with active learning at low label rates. We then focus on the {\it pool-based} active learning paradigm, wherein we iterate between (1) computation of a SSL classifier {\it given the current labeled data} and (2) selection and subsequent labeling of unlabeled query points identified by an acquisition function. We compare our method with various acquisition functions against other standard machine learning methods on the Moving and Stationary Target Acquisition and Recognition (MSTAR) dataset for ATR with small amounts of labeled data. Our main results show that our active learning method for SAR data can outperform state of the art SAR classification methods while using only a fraction of the labeled data used in existing approaches. 

The rest of the paper is organized as follows. In Section \ref{sec:SAR} we overview previous approaches to ATR in SAR imagery. In Section \ref{sec:math} we give the mathematical formulations of graph-based learning and active learning on graphs. In Section \ref{sec:pipeline} we describe our end to end pipeline for constructing graphs with the CNNVAE embedding and applying active learning, while in Section \ref{sec:results} we present our results on the MSTAR dataset. Finally we conclude in Section \ref{sec:conc}.

\subsection{PREVIOUS WORK ON SAR DATA}
\label{sec:SAR}

The previous work on Automatic Target Recognition in SAR imagery is largely focused on the Moving and Stationary Target Acquisition and Recognition (MSTAR) dataset\cite{MSTAR}, which is described in detail in Section \ref{sec:results}. The previous work can be split into pre-deep learning approaches that used hand-tuned features with some basic machine learning methods, like support vector machines (SVM), and deep learning approaches that use convolutional neural networks (CNN).

The pre-deep learning work dates back to the 1990s on using feature extraction with scattering models on SAR imagery \cite{koets1999feature,potter1997attributed}. Subsequently, researchers turned to basic machine learning techniques, including SVM \cite{zhao2001support}, a combination of SVM and Adaboost \cite{wang2006performance}, and other types of adaptive boosting \cite{sun2007adaptive}. Other works have considered filtering SAR imagery (with median filtering) for noise removal to aid in shadow detection \cite{raynal2014shadow}. Finally, hand-tuned covariance descriptor features were used to embed the MSTAR dataset into Euclidean space where SVM and other classical machine learning techniques can be used for classification \cite{dong2014target}.

Deep learning approaches are more recent in the literature. The main difficulty with applying deep learning to SAR imagery, and specifically the MSTAR dataset, is that the dataset is quite small, containing less than  4000 grayscale images. Deep learning normally requires vast amounts of training data to work efficiently and avoid overfitting.  CNNs typically have a number of initial convolutional layers that are are used for feature extraction, followed by a number of fully connected layers that map the features to class predictions. Due to the locality and translation invariance of convolutional layers, the number of parameters per convolutional layer is quite small, and it is often the case that the fully connected layers contribute the vast majority of parameters in a CNN. This has led many researchers to propose modifications to the fully connected part of CNNs when training data is limited.

An early technique involved augmenting a CNN with an initial sparse autoencoder layer that takes random patches of SAR images as input, and reconstructs them under a sparsity constraint \cite{chen2014sar}, presumably to remove noise. The output of the sparse autoencoder was then fed into a CNN without any fully connected layers. The All-Convolutional Networks (A-ConvNets) model was proposed in subsequent works \cite{wang2015application,chen2016target}, which aimed to reduce the model complexity of CNNs by replacing the fully connected layers with convolutional layers in order to prevent overfitting. Another approach adds additional regularization when training the CNN to prevent overfitting. Regularizations include max-norm regularization of convolutional kernels, special initialization methods for network weights, and adapted learning rates for some \emph{priority classes} \cite{wagner2017deep}. Another approach uses feature extractors that are unsupervised, or only mildly supervised. The Euclidean Distance Restricted Autoencoder method\cite{deng2017sar} uses an autoencoder, in which the loss is modified to encourage training points from the same class to have similar latent representations, to extract features from SAR imagery. The classification is then performed on the autoencoder features with linear SVM. Other researchers have considered reducing model complexity by simply using a standard CNN with as few parameters as possible \cite{coman2018deep}.

\section{MATHEMATICAL FORMULATION} \label{sec:al-gbssl}
\label{sec:math}

We now discuss the mathematical formulations of graph-based SSL and active learning that are used in this paper.

\subsection{GRAPH-BASED SEMI-SUPERVISED LEARNING}
Consider an input set of $d$-dimensional feature vectors $\{\bx_1, \bx_2, \ldots, \bx_n\} =: \mcl X \subset \mbb R^d$ and a {\it labeled set} of indices $\mcl L \subset \{1, 2, \ldots, n\}$ that identifies which inputs have observed labels $\{y_j\}_{j \in \mcl L}$.  In ATR, the labels represent classifications of the targets in the images, and so $y_j \in \{1, 2, \ldots K\}$ represents the classification of input $\bx_i$ into one of $K$ classes. These labels can be represented by their corresponding {\it one-hot encodings}, $\tilde{\be}_{y_j}  \in \mbb R^K$, where $\tilde{\be}_k$ is the $k^{th}$ standard basis vector in $\mbb R^K$. Semi-supervised learning (SSL) is the task of inferring labels for the {\it unlabeled set } $\mcl U = \{1,\ldots, n\} - \mcl L$ of data, given the known labels on $\mcl L$.

Graph-based methods for SSL leverage the geometric structure of a similarity graph imposed on the set of feature vectors $\mcl X$ of the inputs to infer the classification of the unlabeled data from the labeled data. We construct a graph $G(\mcl X, W)$ where the nodes represent the inputs $\mcl X$ and the edges are represented by a similarity weight matrix $W \in \mbb R^{n \times n}$, with non-negative, symmetric weights $W_{ij} = W_{ji} \ge 0$ that measure the similarity between node vectors  $\bx_i, \bx_j$. For example, a common similarity measure is the Gaussian kernel $W_{ij} = \exp(-\|\bx_i - \bx_j\|_2^2/\sigma^2)$ with kernel width parameter $\sigma > 0$. The degree matrix $D \in \mbb R^{n \times n}$ of the graph is a diagonal matrix containing the degrees of each node $d_i = \sum_{j = 1}^n W_{ij}$, along the diagonal.

Many graph-based methods for SSL utilize graph Laplacian matrices, such as the graph Laplacian matrix $L = D - W$. Other graph Laplacian matrices are common, such as the {\it symmetric normalized} graph Laplacian $L_n = I - D^{-1/2}W D^{-1/2}$ and the {\it random walk} graph Laplacian $L_r = I - D^{-1}W$. The matrices are positive semi-definite operators with a non-trivial null space; if the graph is connected, then this null-space is one dimensional. For a connected graph we  order the eigenvalues of these matrices as $0 = \lambda_1 < \lambda_2 \le \lambda_3 \le \ldots \le \lambda_n$ with associated real-valued eigenvectors $\bv_1, \ldots, \bv_n \in \mbb R^n$ (See Ref.~\citenum{von_luxburg_tutorial_2007}).

Various graph-based models have been proposed for inferring labels for the unlabeled data, wherein a graph function $u : \mcl X \rightarrow \mbb R^C$ provides information for inferring unknown labels from the labeled nodes ($j \in \mcl L$) according the graph topology\cite{calder_poisson_2020, TPAMI14, zhu_semi-supervised_2003}. Laplace learning is a widely used method in graph-based semi-supervised
learning, originally proposed in Ref.~\citenum{zhu_semi-supervised_2003}, that seeks a graph harmonic function to
extends the labels from the labeled nodes to the unlabeled nodes. More recent methods have been developed that leverage the notion of a graph cut, or graph total variation problem \cite{bertozzi_diffuse_2016,MBC18,MTVC,MBYL17}. For very large datasets and low label rates, Laplace learning can give poor results, and several methods have been proposed that give more stable propgation of labels in this regime, including  $p$-Laplace methods for $p>2$ \cite{flores2022algorithms}, reweighted Laplace  learning \cite{calder2019properly,shi2017weighted}, and Poisson learning \cite{calder_poisson_2020}. In our present work, we will focus on applying Laplace learning once we have created an improved similarity graph via the pipeline described in Section \ref{sec:pipeline}. Future work will focus on improving these techniques with more sophisticated graph learning methods. 

\begin{figure}[t]
    \centering
    \includegraphics[width=0.6\textwidth]{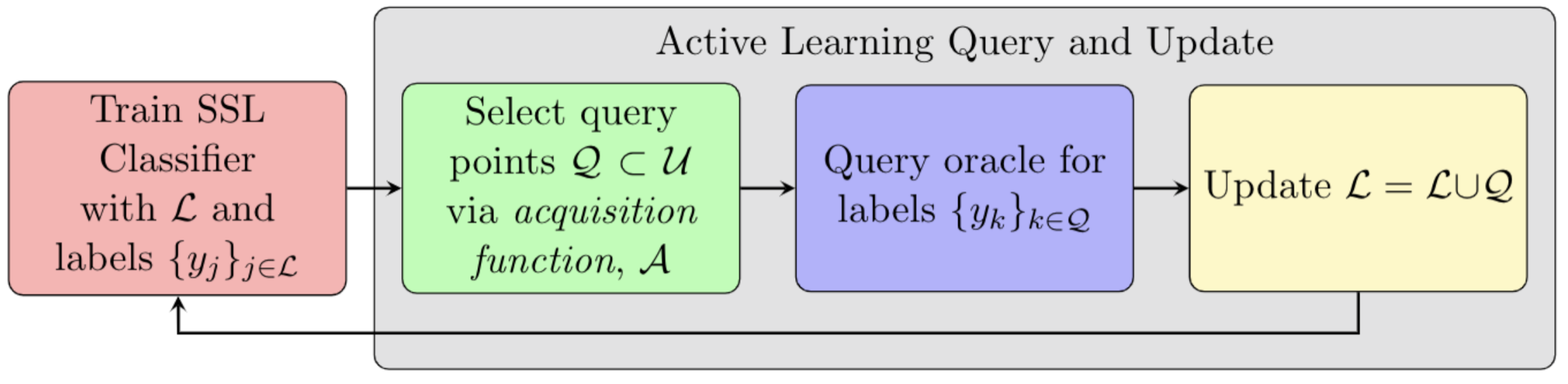}
    \caption{Flowchart for Active Learning Iterations.}
    \label{fig:al-flowchart}
\end{figure}

\subsection{ACTIVE LEARNING}
Active learning takes the next natural step from the semi-supervised learning problem by selecting currently unlabeled points $i \in \mcl U$ to then label via an oracle (i.e., human in the loop) to add to the labeled data and thereby improve the underlying semi-supervised classifier. The active learning process described here iterates between (1) solving for a graph-based semi-supervised classifier given the current labeled data $(\mcl L, \{y_j\}_{j \in \mcl L})$ and (2) identify a {\it query set} $\mcl Q \subset \mcl U$ of unlabeled points to then label and subsequently update the labeled data. The query set $\mcl Q$ is found by using a real-valued {\it acquisition function} $\mcl A : \mcl U \rightarrow \mbb R$ that quantifies how useful it would be to label an individual unlabeled point $\bx_i, i \in \mcl U$. We focus on {\it sequential active learning}, wherein the query set contains only a single point at each iteration $|\mcl Q| = 1$, as opposed to {\it batch active learning} wherein multiple points are selected at each iteration (i.e., $|\mcl Q| > 1$).

Active learning has been applied in various domains, and the associated active learning problem of identifying which unlabeled points would be the ``best'' to label is an active area of research. Similar to other machine learning domains like reinforcement learning, successful active learning requires a proper balance between {\it exploration} and {\it exploitation} of the dataset\cite{miller_model-change_2021}. When few labeled data are available to the classifier, it is desirable for the acquisition function to {\it explore} the extent of clustering structure of the dataset. Once the dataset domain has been properly ``explored'', then it is desired that the acquisition function selects datapoints that {\it exploit} the current classifier's decision boundaries to refine the exact boundaries between classes. Oftentimes, acquisition functions can be viewed as either exploratory or exploitative, though it is desired that an acquisition function automatically balance exploratory behavior first and then exploit. 

Uncertainty sampling\cite{settles_active_2012} is a common acquisition function that chooses to label points of which the current semi-supervised classifier is most uncertain. Variance reduction\cite{settles_active_2012, ji_variance_2012, ma_sigma_2013} selects unlabeled points that would maximally reduce the variance in an associated Bayesian probabilistic model on the current classifier. Model change methods\cite{miller_model-change_2021, karzand_maximin_2020} use the amount by which the current classifier would change as a result of labeling an unlabeled point to select query points. We apply these acquisition functions to the application of interest of ATR on SAR data, and we describe them further in Section \ref{sec:results} where we present our results.



\section{GRAPH-BASED ACTIVE LEARNING PIPELINE} \label{sec:pipeline}
\label{sec:pipeline}

We now describe our novel data pipeline for applying graph-based active learning to SAR data. We utilize neural network architectures called {\it variational autoencoders}\cite{kingma_introduction_2019,kingma2013auto} to learn latent representations of the SAR images from which we construct our similarity graph. Constructing similarity graphs from straightforward Euclidean distances between raw images (including SAR images) fails to capture invariances (e.g., translations and rotations) in the data and is susceptible to noise effects in the data collection process. We apply variational autoencoders with the design of learning better latent representations of the SAR data for constructing the similarity graph which we use to apply graph-based active learning. Our use of variational autoencoders to construct graphs for semi-supervised learning is similar to previous work in graph-based learning \cite{calder_poisson_2020}, except that slightly different network architectures are used in this work.

Variational Autoencoders\cite{kingma_introduction_2019,kingma2013auto} (VAE) transform the input data to (usually) a lower-dimensional space via the use of an ``encoder'' structure from which a ``decoder'' structure is designed to reconstruct the input data. The encoder and decoder neural architectures we use involve convolutional layers, as is natural for input data with spatial relationships like image data with translational and rotational invariances in the similarities. Hence, we refer to our VAE architecture as a CNNVAE (Convolutional Neural Network Variational AutoEncoder).

\begin{figure}
    \centering
    \includegraphics[width=0.7\textwidth]{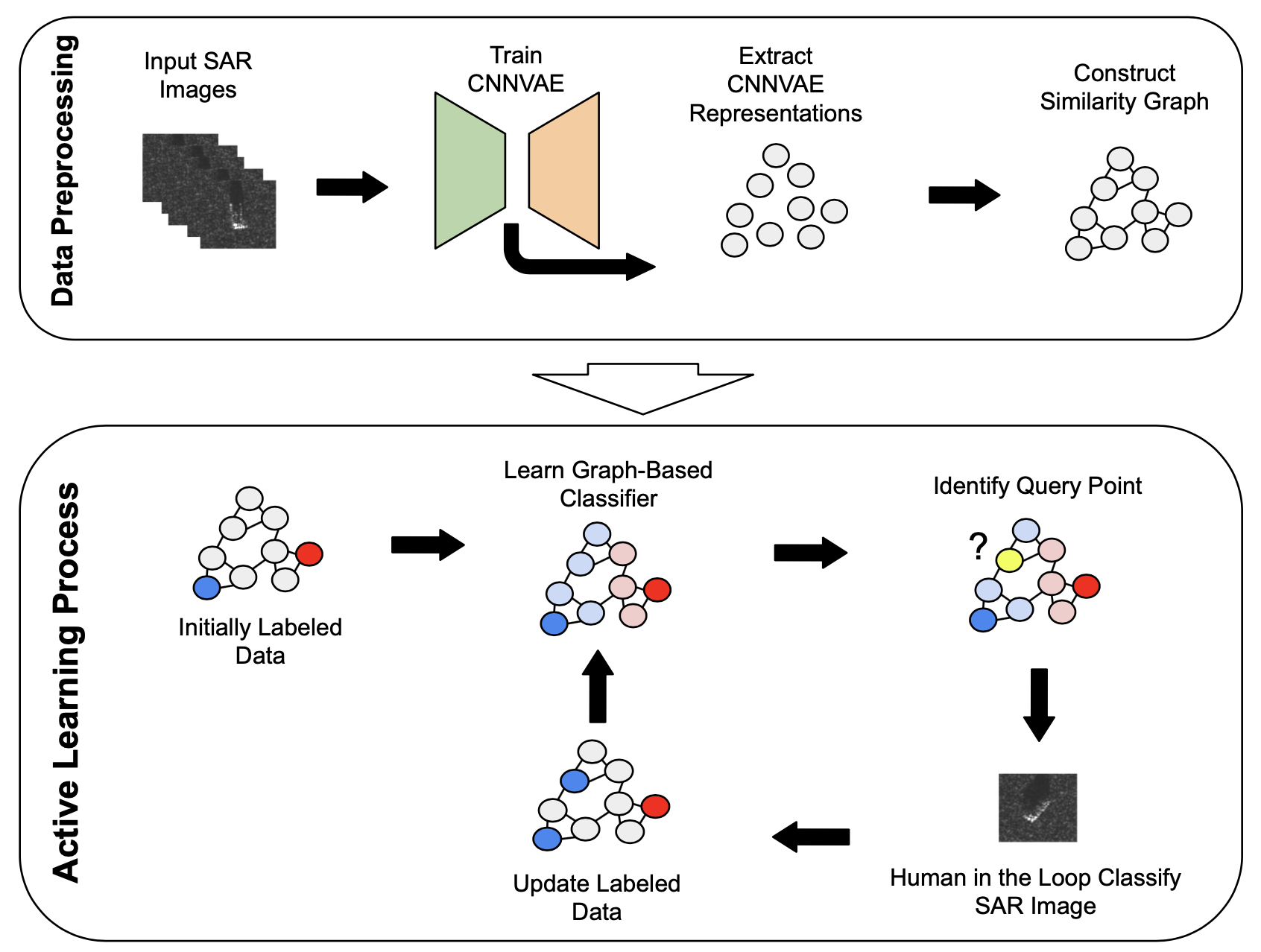}
    \caption{CNNVAE + Graph-Based SSL + Active Learning Pipeline. In the Data Preprocessing phase, we train a CNNVAE on the input SAR images and subsequently construct a similarity graph from the extracted latent representations. In the Active Learning Process phase, we initially label some of the inputs and then proceed to iteratively (1) learn graph-based classifiers and (2) identify and label previously unlabeled data. }
    \label{fig:pipeline}
\end{figure}

Figure \ref{fig:pipeline} shows our pipeline for processing the SAR data. We first train the CNNVAE to learn lower-dimensional embeddings of the SAR imagery data, and then use the learned embeddings to construct a similarity graph on the inputs. This similarity graph is then used for inferring the labels (classifications) of the unlabeled data from the small amount of labeled data that we possess initially. We then apply graph-based active learning acquisition functions in order to sequentially select unlabeled points for a human in the loop then label and add to the labeled data.

\subsection{GRAPH-BASED CLASSIFIERS} \label{subsec:gbssl}

Once we have constructed a similarity graph from the CNNVAE representations of the SAR images, we then apply Laplace learning\cite{zhu_semi-supervised_2003} to infer labels for the unlabeled data from a small set of initially labeled data. For our experiments on the MSTAR dataset, we choose this initial set by selecting uniformly at random a single image from each class. In applications, this reflects the reasonable assumption that we have at least one known example of each possible class for our multi-class classification problem. 

Laplace learning computes a graph harmonic function to extend the labels of the initial set to the rest of the graph. That is, given the labeled set of indices $\mcl L$ with labels $\{y_j\}_{j \in \mcl L}$, then with the corresponding one-hot encodings $\{\tilde{\be}_{y_j}\}_{j\in \mcl L}$, Laplace learning solves for a node function $u : \mcl X \rightarrow \mbb R^C$ that satisfies
\[
\begin{cases}
    u(\bx_i)^T = \frac{1}{d_i}\sum_{j = 1}^n W_{ij} u(\bx_j)^T & \text{ for } i \not\in \mcl L \\
    u(\bx_j) = \tilde{\be}_{y_j} & \text{ for } j \in \mcl L.
\end{cases}
\]
The solution to this system of equations can be written in terms of the graph Laplacian $L = D- W$. Letting $L_{\mcl S, \mcl T}$ denote the submatrix of $L$ whose rows and columns are restricted to the indices in $\mcl S \subset [n]$ and $\mcl T \subset [n]$, respectively. Without loss of generality, we may reorder the indices so that the first $|\mcl L|$ nodes correspond to the labeled nodes in $\mcl L$. Then, the solution $u^\ast$ can be written in matrix form
\[
    U^\ast := \begin{pmatrix} u^\ast(\bx_1)^T \\ u^\ast(\bx_2)^T \\ \vdots \\ u^\ast(\bx_n)^T \end{pmatrix} = \begin{pmatrix} Y \\ -L_{\mcl U, \mcl U} L_{\mcl U, \mcl L} Y \end{pmatrix},
\]
where $Y \in \{0,1\}^{|\mcl L| \times K}$ is the matrix whose rows are the one-hot encodings of the corresponding labeled nodes. By properties of graph harmonic functions, every entry of the solution $U^\ast$ is non-negative and one can identify the resulting classifier as 
\begin{equation}\label{eq:comp-labels}
    y^\ast(i) := \argmax_{k=1, \ldots, K} \ u^\ast(\bx_i)_k.
\end{equation}
Note that on the labeled nodes the classifier indeed recovers the proper classifications, i.e., $y^\ast(j) = y_j$.

Furthermore, one can view the Laplace learning solution $U^\ast$ as the {\it maximum a posteriori (MAP)} estimator of an associated Gaussian Random Field (GRF) over the set of nodes, where the node function on the unlabeled set $U_{\mcl U}$ given the observed labels on $\mcl L$ follows the conditional Gaussian distribution $U_{\mcl U} | U_{\mcl L} = Y \sim \mcl N(U^\ast_{\mcl U}, L_{\mcl U, \mcl U}^{-1})$\footnote{To clarify, we mean that the covariance of each {\it column} of $U_{\mcl U}$ shares the covariance matrix $L_{\mcl U, \mcl U}^{-1}$}. This probabilistic perspective of Laplace learning forms the basis of various graph-based acquisition functions for applying active learning with Laplace learning\cite{zhu_combining_2003}. 

The covariance matrix of the GRF associated with Laplace learning, $C = L_{\mcl U, \mcl U}^{-1}$, provides a useful measure of uncertainty in label inferences across the geometry of the graph, but is prohibitively costly to calculate for larger datasets. One could consider a spectral truncation model for constructing a low-rank approximation to this covariance matrix by using the first $m < n$ eigenvalues and corresponding eigenvectors of the graph Laplacian matrix $L$. However, the resulting covariance matrix is ill-conditioned for updating within active learning iterations. As a result, we use a related graph-based model we term ``Gaussian Regression'' for obtaining a low-rank covariance matrix that is computationally efficient and numerically stable for updating during the active learning iterations. { To be clear, we use Laplace learning for the underlying graph-based semi-supervised classifier, but use the following model for estimating the variance and uncertainty over the unlabeled data an acquisition function calculations in Section \ref{subsec:gb-al}.}

The Gaussian Regression\cite{miller_efficient_2020} (GR) graph-based model is the solution to the following optimization problem:
\begin{align*}
    U_{GR}^\ast &= \argmin_{U \in \mbb R^{n \times K}} \ \langle U, L U \rangle_F + \frac{1}{\gamma^2}\sum_{j \in \mcl L} \|u(j) - \tilde{\be}_{y_j}\|_2^2,
\end{align*}
with the hyperparameter $\gamma > 0$ and where $\langle \cdot, \cdot \rangle_F$ denotes the Frobenius inner product between matrices of compatible dimensions. The solution $U_{GR}^\ast$ can be written in closed form: 
\[
    U_{GR}^\ast = \frac{1}{\gamma^2} \lp L + \frac{1}{\gamma^2}P^T P\rp^{-1}Y =: \frac{1}{\gamma^2}C_{GR}Y,
\]
where the $P \in \mbb R^{|\mcl L| \times n}$ is the projection matrix onto the labeled index set $\mcl L$.
Futhermore, like Laplace learning, the optimizer $U_{GR}^\ast$ can be used to define a classifier as in Equation \ref{eq:comp-labels}. 

The GR model $U^\ast$ also can be identified with the MAP estimator of an associated GRF\cite{bertozzi_posterior_2021} that follows the Gaussian distribution $U \sim \mcl N(U_{GR}^\ast, C_{GR}).$ Similar to the covariance matrix $C = L_{\mcl U, \mcl U}^{-1}$ of Laplace learning, the GR covariance matrix $C_{GR} \in \mbb R^{n \times n}$ is prohibitively costly to compute for use in active learning on larger datasets. We use a low-rank approximation of $C_{GR}$ using the $m < n$ smallest eigenvalues and their corresponding eigenvectors of the graph Laplacian matrix $L$. Let
\[
    \Lambda := \operatorname{diag}(\lambda_1, \lambda_2, \ldots, \lambda_m), \quad V := [\bv_1, \bv_2, \ldots, \bv_m],
\]
and we approximate the covariance matrix by\cite{miller_model-change_2021} 
\begin{equation} \label{eq:gr-cov-approx}
    C_{GR} \approx V \lp \Lambda + \frac{1}{\gamma^2} V^T P^T P V\rp^{-1} V^T =: V \Sigma_{GR} V^T.
\end{equation}
This low-rank approximation only requires the inversion of an $m \times m$ matrix, which size $m < n$ the user is allowed to choose and in practice can be chosen to be magnitudes of size smaller than the size of the dataset, $n$.

\subsection{GRAPH-BASED ACQUISITION FUNCTIONS} \label{subsec:gb-al}

We now discuss the graph-based acquisition functions $\mcl A: \mcl U \rightarrow \mbb R$ for applying in the active learning iterations on the MSTAR dataset in Section \ref{sec:results}. We apply Random Sampling, Uncertainty Sampling, VOpt, Model Change (MC), and a novel acquisition function MCVOpt. Random sampling selects query points by sampling uniformly at random from the unlabeled set at each iteration. Uncertainty Sampling\cite{settles_active_2012} selects query points that the current graph-based classifier is {\it the most uncertain} about at each iteration, as quantified by the following measure of uncertainty: Given the Laplace learning output $u^\ast(\bx_i) \in \mbb R^K$ at the unlabeled node $i \in \mcl U$, calculate the {\it margin} between the first and second largest elements of $u^\ast(\bx_i)$:
\[
    Margin(i) := y^\ast(i) - \lp \argmax_{k\not= y^\ast(i)} \ u^\ast(\bx_i)_k \rp.
\]
One can interpret a {\it smaller} margin to represent a node that has {\it more uncertainty} in the resulting classification from Equation \ref{eq:comp-labels}. There are various uncertainty measures one could apply for uncertainty sampling and our tests showed similar results for each, although the margin calculation was the top performing. To fit into a unified framework of {\it maximizing} acquisition functions to identify query points at each iteration, we therefore write the Uncertainty Sampling acquisition function as
\[
    \mcl A_{U}(i) = 1 - Margin(i).
\]

The VOpt\cite{ji_variance_2012} acquisition function calculates the amount that the trace of the covariance matrix $C = L_{\mcl U, \mcl U}^{-1}$ decreases as a result of adding an unlabeled point $i \in \mcl U$ to the labeled set $\mcl L$. Owing to the computational difficulties of using $C$ or a spectral truncation to approximate $C$, we calculate VOpt using the low-rank approximation of the GR covariance matrix in Equation \ref{eq:gr-cov-approx} of Section \ref{subsec:gbssl}. For an unlabeled point $i \in \mcl U$, the resulting covariance matrix if $i$ were to be added to the labeled set $\mcl L$ can be written as:
\[
    C^{+k}_{GR} \approx V \lp \Lambda + \frac{1}{\gamma^2} V^T P^T P V + \frac{1}{\gamma^2} V^T \be_i \be_i^T V \rp^{-1} V^T = V \lp \Sigma_{GR}^{-1} + \frac{1}{\gamma^2} V^T \be_i \be_i^T V \rp^{-1} V^T.
\]
This allows us to write
\begin{align*}
    Tr\left[ V \lp \Sigma_{GR}^{-1} + \frac{1}{\gamma^2} V^T \be_i \be_i^T V \rp^{-1} V^T \right] &= Tr\left[\lp \Sigma_{GR}^{-1} +  \frac{1}{\gamma^2} V^T \be_i \be_i^T V \rp^{-1} \right] \\
    &= Tr\left[  \Sigma_{GR} \right] - \frac{1}{\gamma^2 + \be_i^T V \Sigma_{GR} V^T \be_i} \left\|\Sigma_{GR}V^T\be_i\right\|_2^2
\end{align*}
by properties of the trace of a matrix, the orthonormality of the eigenvectors of the graph Laplacian\footnote{resulting from the symmtry of $L$ }, and the Woodbury matrix identity. The first term of the final line of the above equation is a constant that is common to the corresponding equation for each unlabeled index $i$. Therefore, our acquisition function for VOpt can be written as
\begin{equation}\label{eq:vopt}
    \mcl A_{V}(i) := \frac{1}{\gamma^2 + \be_i^T V \Sigma_{GR} V^T \be_i} \left\|\Sigma_{GR}V^T\be_i\right\|_2^2,
\end{equation}
which only requires the storage of the $m \times m$ matrix $\Sigma_{GR}$.

Model Change\cite{miller_model-change_2021, miller_efficient_2020} (MC) is a recently proposed graph-based acquisition function that quantifies the amount by which the underlying graph-based model (e.g., $U_{GR}^\ast$) would change {\it if an unlabeled point $i$ were added to $\mcl L$} with the label $y^\ast(i)$ predicted by the current model. Instead of recomputing the Laplace learning for every possible combination of unlabeled point $\bx_i$ with pseudolabel $y^\ast(i)$, we use the spectral truncation approximation of the Gaussian Regression model to efficiently approximate this ``model change'' quantity. The MC acquisition function can be written as
\begin{equation} \label{eq:mc}
    \mcl A_{MC}(i) := \frac{\|u^\ast(\bx_i) - \tilde{\be}_{y^\ast(i)}\|_2}{\gamma^2 + \be_i^T V \Sigma_{GR} V^T \be_i}\left\|\Sigma_{GR}V^T\be_i\right\|_2.
\end{equation}
Noting the similarity between Equations \ref{eq:vopt} and \ref{eq:mc}, we also consider a different acquisition that combines MC and VOpt into the {\it MCVOpt} acquisition function:
\[
    \mcl A_{MCVOpt}(i) := \frac{\|u^\ast(i) - \tilde{\be}_{y^\ast(i)}\|_2}{\gamma^2 + \be_i^T V \Sigma_{GR} V^T \be_i}\left\|\Sigma_{GR}V^T\be_i\right\|_2^2.
\]
We highlight the squaring of the 2-norm quantity that distinguishes this MCVOpt from the MC acquisition function.


\section{RESULTS} \label{sec:results}
\label{sec:results}

We now present our experimental results on the Moving and Stationary Target Acquisition and Recognition (MSTAR) dataset\cite{MSTAR}. The source code to reproduce all our results is available online\footnote{Source Code: \url{https://github.com/jwcalder/MSTAR-Active-Learning}}. The graph-learning algorithms were implemented with the GraphLearning Python package \cite{calder2022graphlearning}.

\subsection{DATASET DESCRIPTION} 

\begin{figure}[t!]
        \centering
        \includegraphics[width=0.49\textwidth]{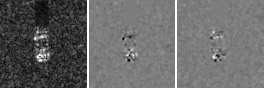}
        \hfill
        \includegraphics[width=0.49\textwidth]{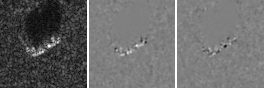}
        \caption[Example SAR images.]{\small Two example SAR images from MSTAR dataset. Pixels in SAR images are complex numbers represented in polar coordinates. For each image, we show (from left to right), the magnitude image, the real part, and imaginary part, of the SAR image. } 
        \label{fig:example-sar}
\end{figure}

The Moving and Stationary Target Acquisition and Recognition (MSTAR) dataset\cite{MSTAR} was collected by
Sandia National Laboratory in a project that was jointly sponsored by the Defense Advanced Research
Projects Agency (DARPA) and the Air Force Research Laboratory (AFRL) in 1998. For an overview tutorial of MSTAR, we refer to \cite{keydel1996mstar}. The dataset contains 6,874 images of 10 types of military vehicles 
(Armored Personnel Carrier: BMP-2, BRDM-2, BTR-60, and BTR-70; Tank: T-62, T-72; Weapon System: 2S1; Air Defense Unit: ZSU-234; Truck: ZIL-131; Bulldozer: D7). A Sandia X-band radar operating at 9.60GHz with a bandwidth of 0.591GHz was used to collect the data; the range
and cross range resolution were both 0.3047m. 
We follow a standard training and testing split according to the angle at which the SAR data was collected; namely, the training data was obtained at an angle of $15\degree$ while the testing data was obtained at $17\degree$. Given this pre-defined train and test split of the data, we accordingly restrict the possible set of active learning query points at any iteration to belong to the training set, and we only test the accuracy on the testing set.

\subsection{PREPROCESSING}
\label{sec:pre}

The original SAR images are of various sizes, and so the magnitude and phase images were all center-cropped to $88\times 88$ pixels. A vast majority ($>99.8\%$) of the pixel values in the magnitude images are within the range $[0,
1]$. The pixels outside this range appear to be noise, and so we clipped the magnitude images to the range $[0,1]$. We then converted each magnitude and phase image-pair into a 3-channel image by taking the first channel to be the magnitude image, and the second two channels to be the real and imaginary parts of the SAR image. More precisely, if $M$ is the magnitude image and $P$ is the phase image, the 3-channel image is given by
\[\left( M, \frac{1}{2}\left( M\cos(P) + 1\right), \frac{1}{2}\left( M\sin(P) + 1\right)\right).\]
This transformation ensures all image channels have pixel values in the range $[0,1]$, which is necessary for the loss function in variational autoencoders. While we include the phase images, we experimented with using only the magnitude images and found very similar results (an accuracy difference of less than 1\%), indicating that the phase images do not contribute a significant amount of information that is not already present in the magnitude images. This finding agrees with previous work \cite{coman2018deep}.

\subsection{CNN ARCHITECTURES}

We trained both fully supervised CNNs and a CNN autoencoder on the 3-channel MSTAR data described in Section \ref{sec:pre}. The fully supervised CNN architecture is quite standard. We used two convolutional layers, with 32 and 64 channels, respectively, with a $2\times 2$ max-pool after the first layer and a $4\times 4$ max-pool after the second. We used ReLU activations in all networks. This results in $6400$ convolutional features that are flattened and fed into a fully connected neural network with two layers and $512$ hidden units. We used dropout\cite{srivastava2014dropout} with rate $0.25$ before the first fully connected layer, and dropout with rate $0.5$ between the two fully connected layers, to help prevent overfitting. We also used a batch normalization\cite{ioffe2015batch} layer betweeeen the two fully connected layers. The usual negative log liklihood classification loss was used, and the network was trained for 50 epochs. 

The CNNVAE architecture is also quite standard, and follows closely the seminal work\cite{kingma2013auto}, with the fully connected layers replaced by convolutional layers. In particular, we used 4 convolutional layers in the encoder, with 8, 16, 32, and 64 channels, respectively. The last layer has a stride of 2 yielding $64$ channels of $11\times 11$ images as output of the encoder. The decoder is symmetrically constructed, except with transposed convolutional layers. The fully connected layers for learning representations had $128$ hidden nodes and the latent dimension is $32$. 

\subsection{GRAPH CONSTRUCTIONS}

In order to apply graph-based learning, we follow previous work \cite{calder_poisson_2020} and build a $k$-nearest neighbor similarity graph on the latent CNN and CNNVAE features.  When we refer to the latent CNN features of either network, we are referring to the output of the initial convolutional layers, once flattened into a vector. For the supervised CNN, the latent CNN features have dimension $6400$, while for the CNNVAE, the latent features have dimension $64\times 11\times 11=7744$. We use an approximate $k$-nearest neighbor search\footnote{We use the Annoy Python package \url{https://github.com/spotify/annoy}.} to efficiently perform the nearest neighbor search in high dimensions. Instead of the usual Euclidean distance, we use the angular metric for the $k$-nearest neighbor search, which is equivalent to normalization all the feature vector to unit norm and using the Euclidean distance. 

After performing a $k$-nearest neighbor search, we construct a self-tuning similarity graph with edge weights
\[w_{ij} = \exp\left( -4|x_i - x_j|^2/d_k(x_i)^2\right),\]
where $x_i$ represents the latent CNN or CNNVAE features of the $i^{\rm th}$ MSTAR image, and $d_k(x_i)$ is the distance in the latent feature space between $x_i$ and its $k^{\rm th}$ nearest neighbor. We used $k=20$ in all experiments and symmetrized the weight matrix by replacing $W$ with $W+W^T$.

\subsection{COMPARISON TO DEEP SUPERVISED LEARNING} \label{subsec:compare-cnn}

\begin{figure}
    \centering
    \subfigure[Accuracy during training]{\includegraphics[clip=true,trim=15 15 16 16,width=0.49\textwidth]{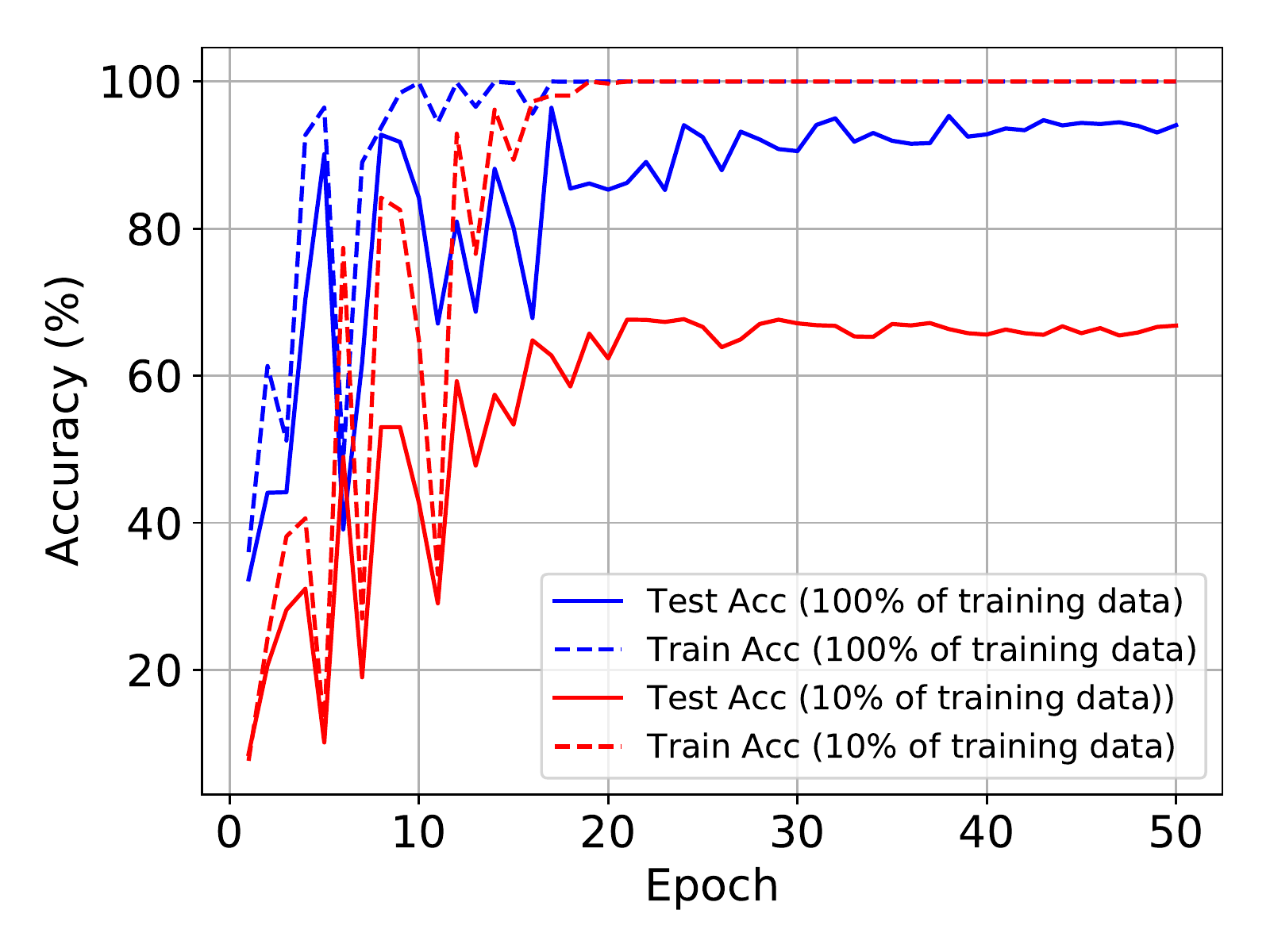}\label{fig:cnn-train}}
    \subfigure[Comparisons to other methods]{\includegraphics[clip=true,trim=15 15 16 16,width=0.49\textwidth]{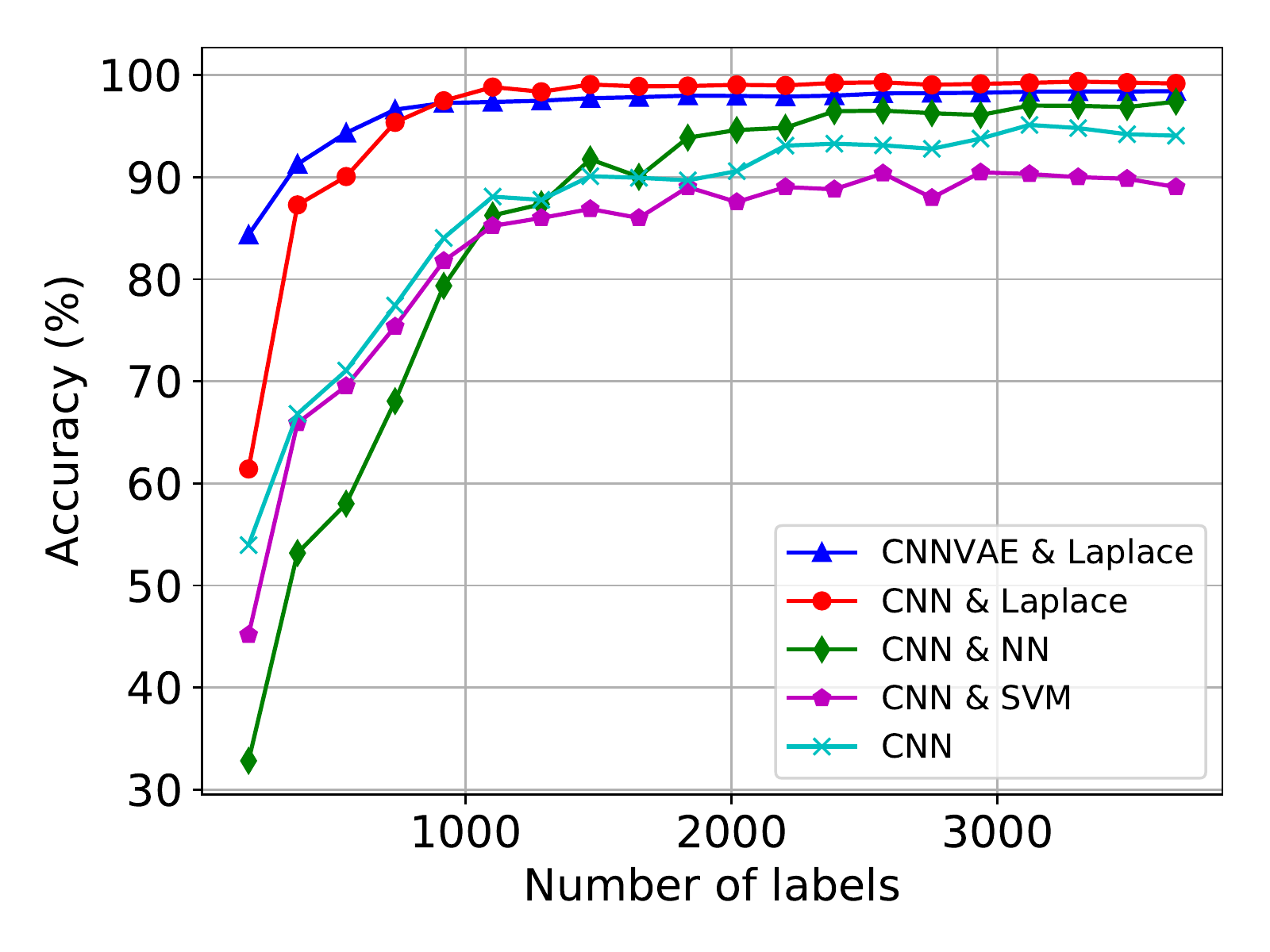}\label{fig:cnn-laplace}}
    \caption{In (a) we show the test and train ATR accuracy over all 50 epochs of training the CNN using all the training data (100\%) as well as using just 10\% of the training data. We clearly see overfitting, especially when only 10\% of the training data is used. In (b), we show a comparison of ATR accuracy on MSTAR for various machine learning algorithms. We trained a CNN on increasing amounts of the training set's data with corresponding ground-truth labels. At each amount of labeled data, the CNN's latent representations were extracted for the {\it entire} MSTAR dataset, and used to evaluate a variety of machine learning algorithms. Applying Laplace learning at each level (CNN \& Laplace) outperformed the other chosen methods; CNN, linear SVM, and nearest neighbors (NN) with $k=5$ neighbors. Furthermore, the accuracies for Laplace learning using representations from the unsupervised CNNVAE are also plotted for comparison. In the regime with small amounts of labeled data (less than 1000 labeled points), graph-based learning with the CNNVAE features is clearly superior to all other methods, while for higher label rates, graph learning with the supervised CNN features performed best. In particular, graph learning with CNN or CNNVAE features always outperformed a fully supervised CNN classifier.}
\end{figure}

We first present results on the efficacy of leveraging both the unsupervised CNNVAE representation learning for improved graph construction and the graph-based semi-supervised learning classifier in our pipeline described in Section \ref{sec:pipeline} for this application of SAR imagery for ATR on the MSTAR dataset. 
We measure the utility of the CNNVAE's learned representations -- which involved no label data -- by comparing against the performance of the representations from a {\it supervised} CNN trained on increasing amounts of labeled data from the predetermined training set.
For each percentage (5\%, 10\%, 15\%, $\ldots$) of labeled data selected uniformly at random without replacement from the predetermined train set, we train a CNN and extract the corresponding latent features for the {\it entire} MSTAR dataset. We train all CNNs for 50 epochs without early stopping. Figure \ref{fig:cnn-train} shows the train and test accuracy over the training epochs using 10\% and 100\% of the training data. We see overfitting in both cases, which is much more extreme in the former case. Looking ahead to Section \ref{subsec:mstar-results}, we can contrast the fully supervised CNN with our active learning results that can obtain better than 99\% accuracy with less than 10\% of the training data. 

After training the fully supervised CNN at different label rates, we then use these CNN representations to evaluate a variety of machine learning algorithms: Support Vector Machine (SVM), nearest neighbors (NN) with $k=5$ neighbors, Laplace learning (where the graph is constructed from these latent features), and the original CNN itself. For example, after training a CNN with $N$ labeled points from the training set and the corresponding latent representations of all 6,784 MSTAR images are extracted from this CNN, then each of the shown machine learning classifiers are trained on those $N$ labeled points (with the exception of Laplace learning which uses a similarity graph defined on the whole dataset). The reported accuracies in Figure \ref{fig:cnn-laplace} are from evaluation on the testing subset of the MSTAR dataset (i.e., those images obtained at an angle of $17\degree$). Applying Laplace learning (CNN \& Laplace, or the red circle curve) at each amount of labeled data clearly outperforms the other chosen methods that use the CNN representations.

Furthermore, we plot -- in the blue triangle curve -- the accuracies for Laplace learning using the {\it unsupervised} CNNVAE representations for comparison. Figure \ref{fig:cnn-laplace} clearly shows that for all levels of labeled data, the graph-based methods achieve superior performance. In the regime with smaller amounts of labeled data (less than 1000 labeled points), the CNNVAE representations actually give a higher accuracy for Laplace learning than the supervised CNN representations. This suggests that the unsupervised representations from the CNNVAE are indeed useful for constructing the similarity graph at the outset of the data processing pipeline where little to no labeled data is available.

From these encouraging results, we proceed in Section \ref{subsec:mstar-results} to introduce active learning into the labeling process to further improve our graph-based pipeline for ATR in the MSTAR dataset.
We briefly note that active learning is arguably most useful for applications in which there is limited available labeled data initially. The inclusion of a ``human in the loop'' (or domain expert) in real-world applications has the greatest impact in the early stages of the active learning process -- the marginal gains in classifier performance by expending the effort to label data is greater when there are few labeled data. Supervised representation learning with a CNN requires a modest to large amount of labeled data in order to learn effective latent representations of the input data, whereas the unsupervised CNNVAE training we propose is accomplished as a preprocessing step without requiring {\it any labeled data} initially. In important applications, the plethora of unlabeled data that is easily produced can be used to straightforwardly train the CNNVAE a single time; the active learning process then allows the practitioner to efficiently choose small amounts of labeled data to yield significant improvements in classification performance without the need for continued retraining of deep learning architectures.

\subsection{MSTAR ACTIVE LEARNING RESULTS} \label{subsec:mstar-results}

We now present our results from applying our novel graph-based active learning pipeline to the ATR task on the MSTAR dataset. As described in Section \ref{sec:pipeline}, we use a CNNVAE to learn a 32-dimensional embedding of the original 88 x 88 images. We then construct a similarity graph from these lower dimensional embeddings, from which we then iteratively select new query points to add to the labeled data; we use Laplace learning as the underlying graph-based classifier for accuracy evaluation. 

\begin{figure}
    \centering
    \includegraphics[width=0.6\textwidth]{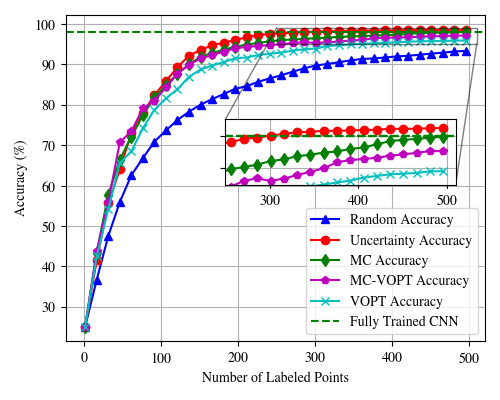}
    \caption{MSTAR Active Learning Results. With initially only one labeled point per class, 500 query points were chosen sequentially according to the shown acquisition functions. MC, MCVOPT, and VOPT used $\gamma = 0.5$ and cutoff of $m = 300$ eigenvalues for the spectral truncated Gaussian Regression computations. Accuracy evaluated according to the Laplace learning classifier.}
    \label{fig:mstar-results}
\end{figure}

Given the similarity graph constructed according to the pipeline described in Section \ref{sec:pipeline}, we perform experiments as follows. A single point per class (i.e., 10 in total) is selected uniformly at random to comprise an initially labeled set. This set is then used as the start for choosing 500 query points for evaluating each acquisition function in the active learning process. After each point is selected and added to the labeled data with its corresponding ground-truth label, the resulting accuracy in the updated Laplace learning model's classifier is recorded. For each acquisition function's set of active learning iterations, an accuracy curve is produced in Figure \ref{fig:mstar-results} by averaging the accuracies over all 10 iterations. These accuracy curves represent how useful the corresponding acquisition functions queries were for improving the underlying classifier's performance in the trials. We use hyperparameter values of $\gamma = 0.5$ and $m = 300$ (spectral truncation cutoff) for the acquisition functions MC, MC-VOPT, and VOPT which utilize the spectral truncated Gaussian Regression model for acquisition function evaluation (Section \ref{subsec:gb-al}).

Since we have ground truth for MSTAR, the active learning queries use this information to provide the new labels.
Each trial used a different random seed to produce the single, initially-labeled point per class. The green dotted line depicts the accuracy for the state-of-the-art CNN method\cite{wagner2017deep, wang2015application, coman2018deep} for ATR on the MSTAR dataset {\it that was trained on the entire training set} (i.e., with 3,671 labeled points). 

With just 300 points labeled during the active learning process, our graph-based semi-supervised classifier achieves the state-of-the-art CNN classifier accuracy! That is, with less than 10\% of the labeled data that are used by modern CNN classifier architectures, we can achieve state-of-the-art classification accuracy on the MSTAR dataset. This is convincing evidence that our proposed graph-based semi-supervised classification and active learning pipeline is a data-efficient and accurate methodology for application to SAR imagery.

By comparing the relative shapes of these accuracy curves, one can measure the relative utility of the given acquisition functions. It is desired that the acquisition function yields the greatest initial increases in accuracy while also ending in the highest overall accuracy. From Figure \ref{fig:mstar-results}, we observe that each of the shown graph-based acquisition functions are superior to random sampling in the earlier stages of the active learning process. They each provide similar initial increases in accuracy for roughly the first 200 choices. Thereafter, while the VOpt, MC, and MCVOpt acquisition functions' corresponding accuracy curves level off at an overall accuracy of around 95\%, Uncertainty Sampling continues to steadily improve to an overall accuracy of just over 97\%. These results suggest that Uncertainty Sampling is the superior choice of acquisition function for graph-based active learning on the MSTAR dataset.

\section{CONCLUSION} \label{sec:conclusion}
\label{sec:conc}

We present a novel machine learning pipeline for active learning with graph-based classification applied to to SAR imagery data. With tests of ATR in the MSTAR dataset, our method proves to be very useful in leveraging small amounts of labeled data to classify the images. Further directions include testing on other SAR imagery datasets as well as improving the graph construction with other unsupervised representation learning methods.

\acknowledgments 
 
This research is funded by an academic grant from the National Geospatial-Intelligence Agency (Award No. \#HM0476-21-1-0003 , Project Title: Graph-based Active Learning for Semi-supervised Classification of SAR Data). Approved for public release, 22-346. Any opinions, findings and conclusions or recommendations expressed in this material are those of the authors and do not necessarily reflect the views of the NGA. 
KM was supported by the DOD's NDSEG Fellowship. JC was supported by NSF DMS-1944925, the Alfred P.~Sloan Foundation, and a McKnight Presidential Fellowship. ALB was supported by DARPA Award number FA8750-18-2-0066 and NSF grant NSF DMS-1952339. 

\bibliography{references} 
\bibliographystyle{spiebib} 

\end{document}